%% file: main.tex
\documentclass[10pt,twocolumn,letterpaper]{article}

\usepackage[pagenumbers]{wacv} 

\input{preamble}

\definecolor{wacvblue}{rgb}{0.21,0.49,0.74}
\usepackage[pagebackref,breaklinks,colorlinks,allcolors=wacvblue]{hyperref}
\usepackage{enumitem}
\usepackage{rotating}

\def\wacvPaperID{220} 
\def\confName{WACV}
\def\confYear{2027}

\title{Statistical Adversaries: Natural Backdoor-like Adversarial Features in Clean Vision Datasets}

\author{
  Paul K. Mandal$^{1,2,3}$ \quad
  Pavan Reddy$^{4}$ \quad
  Tristan Malaty{\'n}ski$^{5}$\\[0.25em]
  {\tt\small paul@research.neurint.ai \quad pavan.reddy@gwmail.gwu.edu \quad tristan@agh.edu.pl}
}

\begin{document}
\maketitle
\footnotetext[1]{Neurint, LLC, Baton Rouge, LA, USA.}
\footnotetext[2]{U.S. Army Cyber Corps, U.S. Army Reserve, USA.}
\footnotetext[3]{Northwestern State University of Louisiana, Natchitoches, LA, USA. \\}
\footnotetext[4]{Automata, Arlington, VA, USA. \\}
\footnotetext[5]{AGH University of Krakow, Krakow, Poland.\\ \\ Correspondence to Paul K. Mandal: {\tt paul@research.neurint.ai}}
\setcounter{footnote}{5}
\input{sec/0_abstract}    
\input{sec/1_intro}
\input{sec/2_related_work}
\input{sec/3_problem_formulation}
\input{sec/4_adversary_construction}
\input{sec/5_experiments}
{
\setlength{\dblfloatsep}{8pt plus 1pt minus 1pt}
\setlength{\dbltextfloatsep}{12pt plus 2pt minus 2pt}
\setlength{\floatsep}{8pt plus 1pt minus 1pt}
\setlength{\textfloatsep}{12pt plus 2pt minus 2pt}
\input{sec/6_results}
\input{sec/7_conclusion}
}

\clearpage
{
    \small

\input{main.bbl}
}

\input{appendix}

\end{document}

%% file: preamble.tex
\usepackage[T1]{fontenc}

\usepackage{tabularx}
\newcommand{\yes}{\ensuremath{\checkmark}}
\newcommand{\no}{\ensuremath{\times}}
\newcommand{\na}{\textit{N/A}}
\newcommand{\mixed}{\(\triangle\)}
\usepackage[normalem]{ulem}
\definecolor{darkamber}{RGB}{145,95,0}

\usepackage{bbm}

\usepackage{placeins}


%% file: sec/0_abstract.tex
\begin{abstract}
Model-specific adversarial attacks have been extensively studied. We study a different failure mode: naturally occurring statistical signals in vision data that can behave as backdoor-like triggers without being maliciously inserted. We call these signals statistical adversaries. We analyse ImageNet to find patterns that are strongly linked to certain labels. We then use statistical controls to remove random correlations from our candidate signals. Finally, we demonstrate that these signals directly and predictably alter model predictions. These statistical adversaries are more targeted than generic corruptions and transfer across different model architectures. This suggests that some vulnerabilities are driven by dataset structure and distribution rather than a single model’s idiosyncrasies. We conclude that ordinary datasets can contain exploitable adversarial surfaces even in the absence of poisoning, and suggest that dataset audits should treat spurious structure not only as a source of bias or interpretability failure, but also as a latent attack surface for vision models.
\end{abstract}

%% file: sec/1_intro.tex
\section{Introduction}
\label{sec:intro}

Although modern vision models achieve strong benchmark performance, these metrics do not guarantee that the features they use are semantically meaningful. Models often use spurious patterns for predictions; these models perform well on benchmarks, but fail to transfer over to real-world scenarios \cite{Geirhosetal20}. Ilyas \etal \cite{NEURIPS2019_e2c420d9} further argue that adversarial examples are features learned from these poorly generalized, spurious patterns. Additionally, both natural and model audits have shown that ImageNet contains harmful spurious features, class-associated frequency shortcuts, and systematic failure cases \cite{Hendrycks_2021_CVPR,Neuhaus_2023_ICCV,Wang_2025_CVPR}. These findings suggest that ordinary, unpoisoned datasets may contain statistical structure that can be leveraged to cause model failures.

Moosavi-Dezfooli \etal \cite{Moosavi-Dezfooli_2017_CVPR} demonstrated that a single adversarial direction can affect many images and transfer across different model architectures. Other work also shows that incorporating information about a target-class distribution can improve cross-model attacks \cite{Naseer_2021_ICCV}. These adversarial directions, however, are generally obtained through optimization against a single victim or surrogate model.

Similarly, frequency-domain attacks use structured frequency components to efficiently search for adversarial perturbations \cite{pmlr-v97-guo19a}. These adversarial directions are identified through responses from the attacked model. Information-geometric attacks similarly derive adversarial directions from Fisher geometry from a trained neural network \cite{Zhao_2019_AAAI}. Therefore, while prior work established that numerous different types of attack directions exist, these directions had to be identified through model-mediated signals.

Our paper seeks to answer what these prior papers do not: can target-specific failure directions instead be constructed from the statistics of the original source dataset, without optimizing a victim or surrogate attack objective? Establishing these directions would connect dataset-level statistics to model sensitivities that are shared across multiple different architectures. We refer to these source-data-derived directions that induce target-specific failures as \textbf{statistical adversaries}. Our study is focused on three major questions:

\begin{enumerate}[label=\textbf{RQ\arabic*.}, leftmargin=*, topsep=0.5em, itemsep=0.25em]
    \item Do models trained on ordinary, unpoisoned data exhibit sensitivity to adversarial directions derived only from the dataset?
    \item Can these directions be constructed without victim-model gradients, queries to a model, or surrogate attack objectives?
    \item Can the resulting directions induce target-specific failures on architectures excluded from candidate development?
\end{enumerate}

To answer these questions, we construct bounded perturbation directions solely from class-conditional ImageNet statistics and evaluate them on held-out, target-negative images across CNN and transformer classifiers, including RegNet-Y-8GF and PVT-v2-B2 after the candidate panel and analysis protocol are frozen. We compare raw class-mean directions against more controlled band-pass diagonal-whitened and Hellinger-motivated frequency-statistical directions, using random, low-pass, spectrum-matched, global-mean, and wrong-target controls. We find that selected source-statistical directions increase evidence for absent target classes beyond repeated matched controls. The held-out evaluation is positive in 22 of 22 model--candidate cells, and an expanded 100-draw spectrum null leaves a corrected residual in 22 of 44 original cells. These effects primarily appear as calibrated false-positive inflation, rank movement, and top-\(k\) entry rather than consistent top-1 target takeover.


Our contributions are:
\begin{itemize}[leftmargin=*, itemsep=0.25em, topsep=0.5em]
    \item We introduce and define \textit{statistical adversaries}: directional perturbations derived only from the training data, which we evaluate through a held-out protocol with norm, frequency, and spectrum-matched controls.
    \item We propose source-only direction families based on band-pass diagonal-whitened class contrasts and Hellinger-motivated frequency statistics, requiring no victim-model gradients, queries, or surrogate attack optimization.
    \item We show that selected source-statistical directions induce target-specific false-positive inflation, rank movement, and top-\(k\) entry across CNN and transformer classifiers, including positive movement in all 22 evaluations on two prospectively held-out architectures, while raw class-mean directions provide a negative baseline.
\end{itemize}

%% file: sec/2_related_work.tex
\section{\texorpdfstring{Related Work}{Related Work}}
\label{sec:related_work}

\begin{figure*}[t]
    \centering
    \includegraphics[width=0.92\textwidth]{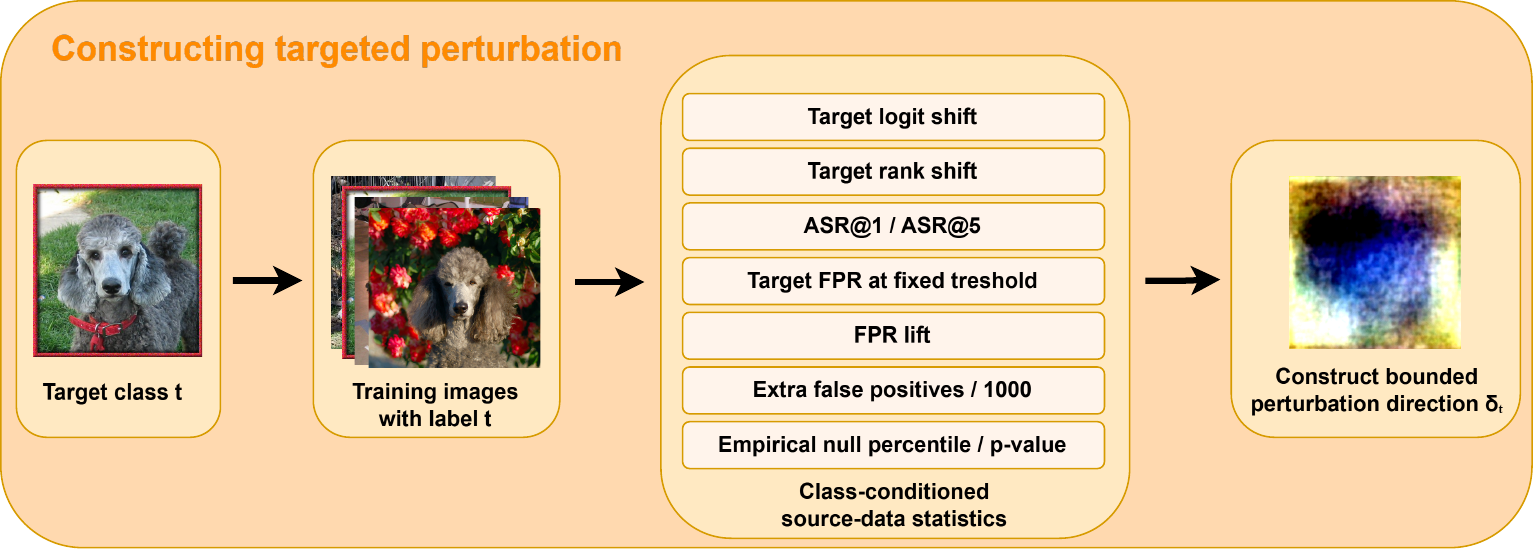}
    \caption{Source-statistics overview. The figure summarizes the dataset-level statistics used to form class-conditioned directions.}
    \label{fig:app-data-statistics}
\end{figure*}

\subsection{Adversarial Attacks}
The current literature on adversarial attacks comes from the observation that imperceptible perturbations can significantly alter the predictions of neural networks. Szegedy et al. were the first to demonstrate the existence of adversarial attacks \cite{Szegedy2014}. Goodfellow et al. later found that vulnerability to adversarial attacks results from local linearity in multidimensional models, and introduced the Fast Gradient Sign Method (FGSM) \cite{Goodfellow2015}; however, Projected Gradient Descent (PGD) is considered the strongest type of attack and serves as the benchmark for evaluating model robustness \cite{madry2018towards}. 

More relevant to our evaluation settings are studies showing that a single perturbation can effectively attack a significant portion of the dataset for a given model \cite{Moosavi-Dezfooli_2017_CVPR}, and in some cases even across different architectures by taking into account information about the target class distribution \cite{Naseer_2021_ICCV}, by adjusting the statistics of intermediate features of a given class instead of directly attacking the logits \cite{Inkawhich_2019_CVPR} or by correlating decision boundaries with geometric properties \cite{Khrulkov2018}. Black-box attacks from a frequency-domain perspective \cite{pmlr-v97-guo19a, Duan_2025_ICIC} and attacks based on information geometry from Fisher's geometry of a trained network \cite{Zhao_2019_AAAI}, further demonstrate that the \emph{shape} of an effective perturbation is constrained by a structure that is not specific to any particular model. Nevertheless, these approaches determine perturbations using information from the model. In contrast, we explore whether this step can be completely eliminated and the direction determined solely based on the dataset.

\subsection{Decision-Making Features of Neural Networks}
Deep learning classifiers tend to focus on features that differ from human perception and are not related to the target class, a behavior that was observed long before the current era of artificial intelligence \cite{Torralba2011_CVPR}. The best-known example is the recognition of wolves in snowy scenes \cite{Geirhosetal20} using local texture statistics instead of a global shape representation \cite{Hermann2019TheOA, Shankar2020, Suhail2026}. Similar examples occur also on ImageNet, where the classifier learned e.g. to recognize birds based on the background \cite{Xiao_2021_ICLR}. Attributes characterized by imbalance between groups may dominate predictions for minority subgroups \cite{Sagawa2020}. Large-scale audits have revealed the existence of thousands of spurious features—correlated with classes and recognizable by humans—directly within the ImageNet dataset \cite {Singla_2022_ICLR, Neuhaus_2023_ICCV, Beyer2020, Hendrycks_2021_CVPR}, such as by adapting to different frequency ranges that may misdefine a class \cite{Wang_2025_CVPR, Yin_2019_NeurIPS, Wang_2020_CVPR}. SilverLining showed that both spatial and spectral shortcuts can simultaneously emerge in real datasets and that removing such cues requires explicitly controlling frequency-domain confounders \cite{Unnikrishnan_2026_WACV}. These audits are framed as issues related to bias and interpretability: they explain why the model fails in the case of an atypical example, but they do not show that the same spurious structure can be transformed into a controlled, repeatable perturbation of the data able to fool the model. This literature points to the frequency domain as a natural place to explore class-conditional structures and directly provides information on the statistics we use in our work to investigate potential gaps in the ImageNet dataset. Most work on shortcut learning focuses on analyzing failures under distribution shifts. We, however, investigate whether such statistical regularities themselves can serve as perturbation directions.

\subsection{Backdoor Attacks and Natural Triggers}

Backdoor attacks involve introducing triggering patterns during training to cause the model’s predictions to shift in the direction chosen by the attacker. BadNets was the first to demonstrate the effectiveness of backdoor poisoning \cite{Gu2017}. Subsequent work introduced “clean-label” attacks \cite{Turner2019}, hidden triggers \cite{Saha2020}, and semantic backdoors \cite{Liu_2020_ECCV}. However, poisoned examples leave a detectable spectral trace in feature space \cite{Tran_2018_NeurIPS}, although that study uses them to detect an injected trigger rather than to rule out naturally occurring triggers, which is the focus of our research. Ordinary, naturally photographed objects can act as triggers when associated with a target label \cite{Wenger_2021_CVPR,wenger2022naturally}, and naturally occurring reflection patterns on glass or water can serve the same function \cite{Liu_2020_ECCV}. Han et al.~\cite{Han2024NaturalTriggers} study natural semantic backdoors for face-synthesis detection, including a custom-trigger variant derived from long-tailed facial attributes without detector supervision. Unlike our setting, however, their triggers are implanted through generated poisoned training examples, whereas our perturbations are constructed from source-data statistics and applied at test time to independently trained clean models. The most important findings come from \cite{Tao_2022_arXiv, 10918226}, showing that normally trained, unpoisoned models already contain input transformations resembling triggers, which the authors refer to as \emph{natural backdoors}. Our work shares the main thesis of this article, but differs in terms of mechanism and evaluation: instead of reverse-engineering a trigger for a specific, trained model, we derive a potential direction based solely on source-class statistics, without access to any victim model, and then verify whether this single, model-independent direction reproduces backdoor-like behavior across several held-out architectures at once.

The literature on dataset bias shows that ImageNet exhibits a structure that can be exploited; the literature on frequency bias narrows down the range in which this structure typically occurs; the literature on transfer attacks shows that a single direction can generalize to different models; and the literature on natural backdoors shows that models not subjected to poisoning may already contain vulnerabilities resembling triggers. To the best of our knowledge, no previous work combines all four observations into a single process that (i) derives a potential direction exclusively from class-conditional source statistics, (ii) rules out trivial explanations for this effect using norm-matched, frequency-matched, and wrong-target controls rather than only a random baseline, and (iii) evaluates the remaining directions without fine-tuning or access to queries, on architectures where that direction was never built or tested during search. This is a gap that Statistical Adversaries are intended to fill.

\begin{figure*}[t]
    \centering
    \includegraphics[width=0.92\textwidth]{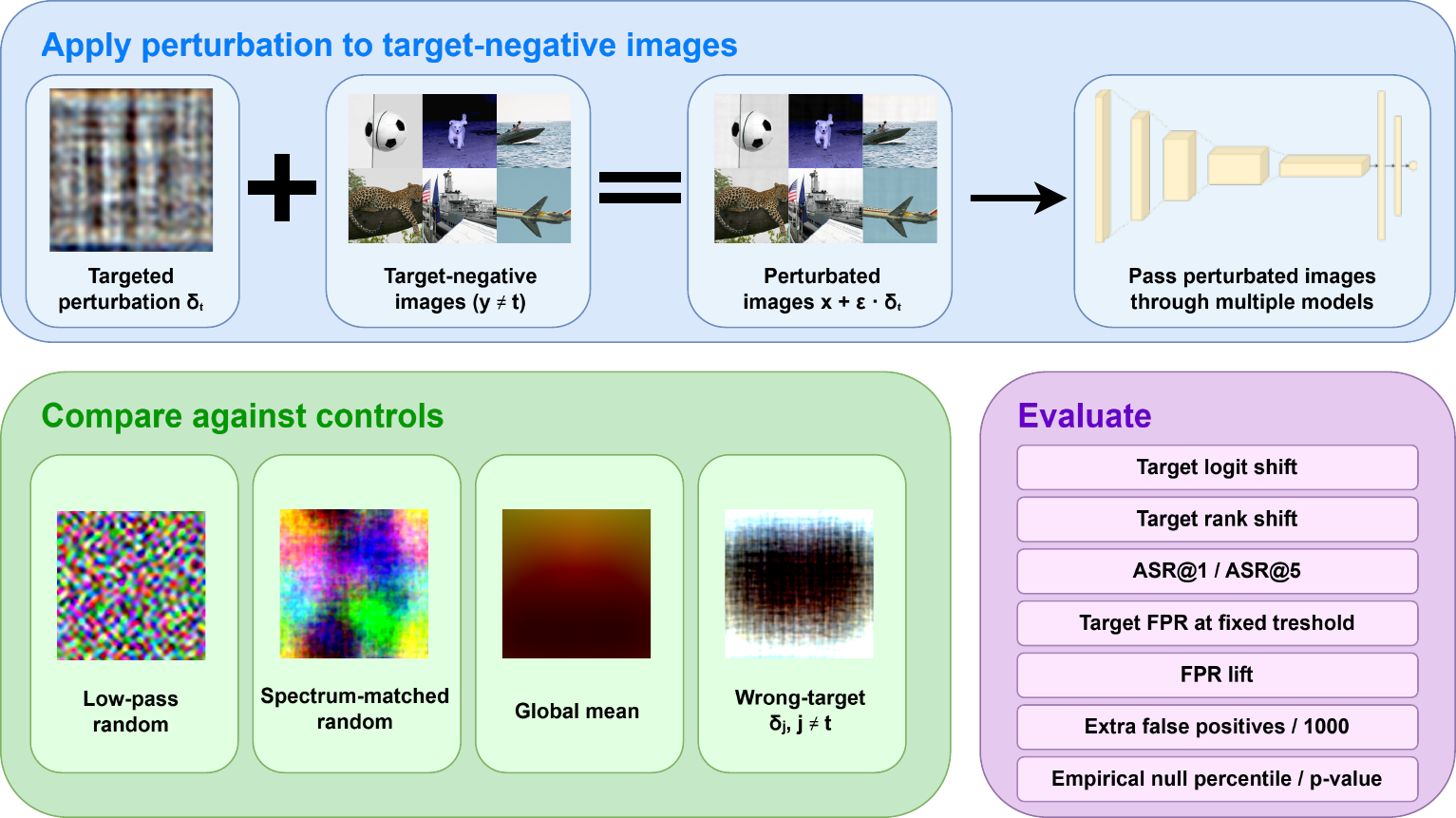}
    \caption{Methodology overview. The figure illustrates the construction pipeline for source-statistical adversary directions.}
    \label{fig:app-methods}
\end{figure*}

%% file: sec/3_problem_formulation.tex
\section{Problem Formulation}
\label{sec:problem}

We construct \emph{universal targeted perturbations from training-set statistics alone}: a single perturbation is built per target class with no access to the victim models at construction time, and is then evaluated for its effect on a family of pretrained classifiers. Figure~\ref{fig:app-data-statistics} summarizes the source statistics used to build these directions.
\subsection{Setup}
\label{subsec:setup}

\paragraph{Images and labels.}
Images are $\boldsymbol{x}\in\mathcal{X}=[0,1]^{C\times H\times W}$ with $C=3$ channels and $H=W=224$, giving flattened dimension $d=CHW$. Class labels are drawn from $\mathcal{Y}=\{1,\dots,K\}$ with $K=1000$.

\paragraph{Source dataset.}
Perturbations are functions of a source training set $\mathcal{D}_{\mathrm{tr}}=\{(\boldsymbol{x}_i,y_i)\}_{i=1}^{N}$ alone; we write $\mathcal{D}_{\mathrm{tr}}^{c}$ for its class-$c$ subset and the number of training samples $n_c=\lvert\mathcal{D}_{\mathrm{tr}}^{c}\rvert$. Evaluation uses a disjoint validation set $\mathcal{D}_{\mathrm{val}}$.

\paragraph{Victim models.}
We attack a fixed family of $M$ pretrained classifiers $\{f^{(m)}\}_{m=1}^{M}$, where each $f^{(m)}\colon\mathcal{X}\to\mathbb{R}^{K}$ returns a vector of $K$ class logits. For each target $t$, evaluation is restricted to the \emph{target-negative} validation images
\begin{equation}
  \mathcal{V}_t=
  \{\boldsymbol{x}\in\mathcal{D}_{\mathrm{val}}:y(\boldsymbol{x})\neq t\},
\end{equation}
so movement toward $t$ is an induced false positive rather than a correct prediction.

\paragraph{Perturbation model.}
For each target $t$ we build a single perturbation $\boldsymbol{\delta}_t\in\mathbb{R}^{d}$, apply it additively, and clamp the result to the image range under an $\ell_\infty$ budget $\varepsilon$:
\begin{align}
  \tilde{\boldsymbol{x}}
    &= \operatorname{clip}_{[0,1]}(\boldsymbol{x}+\boldsymbol{\delta}_t), \\
  \lVert\boldsymbol{\delta}_t\rVert_\infty
    &\le\varepsilon,\qquad
      \varepsilon\in\{8,16,32\}/255 .
  \label{eq:budget}
\end{align}
Construction is \emph{model-free}: $\boldsymbol{\delta}_t=\mathcal{A}(\mathcal{D}_{\mathrm{tr}},t)$ with $\partial\mathcal{A}/\partial f^{(m)}\equiv0$, so it uses no gradients and issues no queries to any $f^{(m)}$. The same $\boldsymbol{\delta}_t$ is reused across every image in $\mathcal{V}_t$ and across all $M$ models, making the attack universal in both inputs and models.

\paragraph{Threat model and attack vector.}
We assume an attacker can modify an image before it reaches the classifier and has access to the source dataset or corresponding class statistics. The target-class direction is computed once offline, so producing a modified image requires only a single addition-and-clipping operation, with no per-image optimization, gradient computation, or pre-submission access to the deployed model. The resulting attack is immediately deployable and opportunistic: an attacker who can choose among available images or submit multiple candidate inputs needs only one modified image to cross the target-specific operating threshold; uniform success across the input population is not required. Where a single false positive can trigger routing, blocking, moderation, retrieval, access-control, or human escalation, the observed FPR inflation can create a high-impact failure surface; transfer to architectures excluded from candidate development further reduces dependence on knowing the deployed architecture.

\subsection{Evaluation metrics}
\label{subsec:metrics}

Our headline metric treats each $f^{(m)}_t$ as a one-vs-rest detector for class $t$. On the clean target-negative images we fix a threshold $\tau^{(m)}_t$ at the $(1-\alpha)$ empirical quantile of the clean target logits, with $\alpha=0.05$, and then hold that threshold fixed after perturbation.\ The resulting false-positive-rate shift is
\begin{align}
  \Delta\mathrm{FPR}_\alpha
    &= \mathbb{E}_{\mathcal{V}_t}
      \mathbbm{1}\{f^{(m)}_t(\tilde{\boldsymbol{x}})>\tau^{(m)}_t\}
      -\alpha, \\
  \mathrm{lift}
    &= 1+\Delta\mathrm{FPR}_\alpha/\alpha .
\end{align}
We report $\Delta\mathrm{FPR}_\alpha$ in percentage points and as extra false fires per 1,000 images. Secondary metrics record the target-logit shift, the target-rank improvement, and targeted ASR@1/5. Target-rank and ASR definitions are provided in Appendix~\ref{app:secondary-metrics}, and inference rules are provided in Appendix~\ref{app:protocol}.

%% file: sec/4_adversary_construction.tex
\section{Source-Statistical Adversary Construction}
\label{sec:construction}

Every perturbation is produced as follows: (i)~form a target-specific \emph{direction} from first- and second-moment statistics of $\mathcal{D}_{\mathrm{tr}}$; (ii)~optionally reshape it with a linear \emph{operator} $\mathcal{O}$ that retains only part of the signal; and (iii)~project the result to the $\ell_\infty$ budget. Figure~\ref{fig:app-methods} illustrates the full construction pipeline.

\subsection{Source-statistical construction}
\label{subsec:directions}

\paragraph{Class means.}
From the class subsets of $\mathcal{D}_{\mathrm{tr}}$ we form the class, global, and complement means
\begin{align}
  \boldsymbol{\mu}_c
    &=\frac{1}{n_c}\sum_{\boldsymbol{x}\in\mathcal{D}_{\mathrm{tr}}^{c}}\boldsymbol{x},
  &
  \boldsymbol{\mu}
    &=\frac{1}{N}\sum_c n_c\boldsymbol{\mu}_c, \\
  \boldsymbol{\mu}_{\bar{t}}
    &=\frac{1}{N-n_t}\sum_{c\neq t}n_c\boldsymbol{\mu}_c .
\end{align}
The raw \emph{class-contrast} direction is the difference between the target and non-target means,
\begin{equation}
  \boldsymbol{g}_t=\boldsymbol{\mu}_t-\boldsymbol{\mu}_{\bar{t}}.
\end{equation}
To prevent a few high-variance coordinates from dominating, we form the \emph{diagonally whitened} contrast
\begin{equation}
  \boldsymbol{w}_t=\boldsymbol{g}_t\oslash(\boldsymbol{s}+\eta),
  \qquad
  s_j=\Bigl(\tfrac{1}{N}\textstyle\sum_i(x_{i,j}-\mu_j)^2\Bigr)^{1/2},
  \label{eq:whiten}
\end{equation}
where $\oslash$ denotes coordinatewise division, $\boldsymbol{s}$ is the vector of per-coordinate standard deviations over $\mathcal{D}_{\mathrm{tr}}$, and $\eta>0$ stabilizes low-variance coordinates.

\paragraph{Operator family and projection.}
A perturbation family is defined by applying an operator $\mathcal{O}$ to either $\boldsymbol{g}_t$ or $\boldsymbol{w}_t$. The final perturbation is obtained by scaling the resulting direction $\boldsymbol{v}_t$ to the budget,
\begin{equation}
  \boldsymbol{\delta}_t
    = \Pi_\varepsilon(\boldsymbol{v}_t)
    = \varepsilon\,\frac{\boldsymbol{v}_t}{\lVert\boldsymbol{v}_t\rVert_\infty},
  \qquad
  \lVert\boldsymbol{\delta}_t\rVert_\infty=\varepsilon .
  \label{eq:proj}
\end{equation}
The confirmation experiments focus on FFT-Hellinger and bandpass diagonal-whitened directions. Appendix~\ref{app:operator-details} gives the full operator definitions and the complete perturbation-family table.

%% file: sec/5_experiments.tex
\section{Experimental Protocol}
\label{sec:experiments}

We test whether class-conditioned statistics estimated from ImageNet can be converted into universal perturbations that increase false positives for a chosen target class. The perturbations are constructed without using victim-model gradients, model-specific optimization, or image-specific updates. Our primary outcome is target-specific false-positive-rate (FPR) inflation. We report targeted ASR@1 and ASR@5 as secondary outcomes.

To determine whether an observed effect is specific to the proposed class-conditioned direction, we compare it with five controls: Gaussian-random noise, lowpass-random noise, spectrum-random noise, the global ImageNet mean, and a direction constructed for the wrong target. Each control is evaluated on the same images and under the same perturbation budget as the corresponding proposed direction. The perturbation budget limits the largest change made to any individual RGB channel value.

\subsection{Data and Evaluation}

\paragraph{Source-statistic construction set.}
We compute the source statistics from all 1,281,167 images in the ImageNet-1K training split. The split contains 1,000 classes, with between 732 and 1,300 training images per class. We use the training split only to estimate the statistics from which the perturbations are constructed. We do not train or fine-tune any of the victim models, and we do not compute FPR, ASR, or attack-success results on the training images.

ImageNet-1K images do not have consistent dimensions. Thus, we resized each image using bicubic interpolation such that its shorter side was 256 pixels. We then center-cropped the image to $224\times224$ and converted to an RGB tensor with values in $[0,1]$. From these images, we calculate per-class and global mean images, first and second moments at each pixel location and color channel, and summaries of how image energy is distributed across spatial frequencies.

\paragraph{Candidate selection and validation sets.}
The ImageNet-1K validation split contains 50,000 images, with exactly 50 images from each class. Candidate discovery requires searching over target classes, perturbation constructions, and perturbation budgets. Evaluating the final candidates on the same images used to select them would therefore overestimate their performance.

Thus, we shuffle the 50 validation images within each class using seed 17 and divide the resulting ordering into non-overlapping slices. Positions 0--4 form the concept-check slice, positions 5--14 form the candidate-validation slice, and positions 15--24 form the confirmation slice. The candidate panel is fixed before the confirmation slice is evaluated, and only results from this slice are used for the headline confirmation analysis. The staged search evaluates 100 targets and 800 target--construction--budget configurations in the broad screen, 8 targets and 64 configurations in concept check, and 4 targets and 16 configurations in candidate validation. Nine validation configurations and two continuity configurations reintroduced from concept check form the final 11-candidate panel.

\paragraph{Frozen confirmation panel.}
The confirmation panel contains 11 target--construction--budget combinations evaluated on the four development models, giving 44 model-level evaluations. Twenty-four correspond to FFT-Hellinger and twenty correspond to bandpass diagonal-whitened mean. The same frozen candidates are subsequently evaluated without reselection on RegNet-Y-8GF and PVT-v2-B2, giving 22 held-out model--candidate cells. Since the panel was assembled during the development stages, the reported results summarize these selected candidates rather than every ImageNet target.

\paragraph{Statistical tests and Benjamini--Hochberg correction.}
We test paired target-threshold crossings with a one-sided exact binomial test, equivalently a one-sided exact McNemar test, and apply Benjamini--Hochberg correction across the 44 confirmation cells. The corresponding held-out tests are corrected separately across the 22 held-out cells; empirical Gaussian and spectrum-control tests are likewise corrected within their prespecified original-model and held-out-model families. This testing protocol is detailed more thoroughly in Appendix~\ref{app:protocol-details}.

\subsection{Victim Models and Controls}

\paragraph{Victim models.}
We evaluate six pretrained ImageNet classifiers: ResNet-50, ConvNeXt-Tiny, ViT-B/16, and Swin-T, followed by the prospectively held-out RegNet-Y-8GF and PVT-v2-B2. The development models and RegNet use pretrained Torchvision weights; PVT-v2-B2 uses pretrained timm weights. The four development models participate in candidate selection, whereas the two held-out models are evaluated only after the 11-candidate panel and analysis protocol are frozen. Their parameters remain frozen, and inference is performed in evaluation mode. The same target-specific perturbation is evaluated on every model.

\paragraph{Perturbation application}
We add our perturbations in the RGB image space prior to ImageNet normalization. After adding the perturbation, pixel values are clipped to the valid RGB range of $[0,1]$. We then apply standard ImageNet normalization. Our $\ell_\infty$ perturbation budgets are $8/255$ and $16/255$. No single RGB value for a given pixel may change more than the perturbation budget.

\paragraph{Controls.}
Each proposed direction is compared with controls evaluated on the same target class, target-negative images, victim model, and perturbation budget. The Gaussian-random control uses 100 random pixel-space noise directions. The lowpass-random and spectrum-random controls use 30 and 100 seeds, respectively, on the four development models. The held-out models use 100 Gaussian-random seeds; lowpass-random and spectrum-random are not rerun on the held-out models. The global-mean and wrong-target controls are deterministic.

%% file: sec/6_results.tex
\begin{figure*}[t]
\centering
\includegraphics[width=0.97\textwidth]{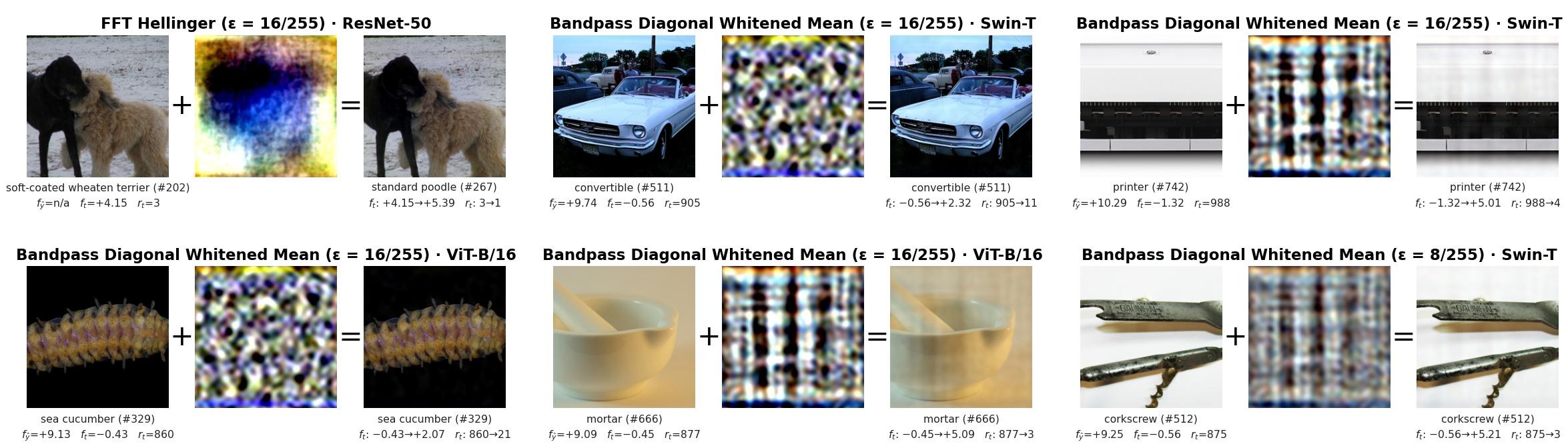}
\caption{Representative confirmation-slice examples. Each triplet shows a clean target-negative image, a visualization of the corresponding source-statistical perturbation, and the perturbed image after applying the perturbation. Panel titles report the construction, perturbation budget, and victim model; text below the images reports the target class, clean and perturbed target logit, and clean and perturbed target rank.}
\label{fig:qualitative-results}
\end{figure*}

\begin{table*}[t]
\centering
\small
\setlength{\tabcolsep}{3.6pt}
\caption{Target-specific false-positive inflation on the frozen confirmation panel. Counts are condition-level prediction decisions rather than unique images. Corrected counts are after Benjamini--Hochberg correction.}
\label{tab:headline-fpr}
\begin{tabular}{@{}lrrrrrrrr@{}}
\toprule
Construction & Cells & Evals. & Clean & Perturbed & FPR & Net add. & Extra FPs & Corrected \\
             &       &        & FPR   & FPR       & lift & FPs      & /1,000    & $q\leq.05$ \\
\midrule
All proposed & 44 & 439,560 & 5.005\% & \textbf{9.689\%} & 1.94$\times$ & \textbf{20,589} & \textbf{46.84} & \textbf{40/44} \\
FFT-Hellinger & 24 & 239,760 & 5.005\% & 7.994\% & 1.60$\times$ & 7,167 & 29.89 & 21/24 \\
Bandpass-whitened & 20 & 199,800 & 5.005\% & 11.723\% & 2.34$\times$ & 13,422 & 67.18 & 19/20 \\
\bottomrule
\end{tabular}
\end{table*}

\begin{table*}[t]
\centering
\small
\setlength{\tabcolsep}{4pt}
\caption{Comparison with clean inputs and matched controls on the original four-model panel. Stochastic control FPRs are averaged over seeds within each model--candidate cell. Corrected counts are after Benjamini--Hochberg correction.}
\label{tab:control-comparison}
\begin{tabular}{@{}lrrrrrr@{}}
\toprule
Comparator & FPR & $\Delta$FPR & Prop.-comp. & Prop. $>$ comp. & Raw $p\leq.05$ & Corrected $q\leq.05$ \\
\midrule
Clean & 5.005\% & 0.000 pp & \textbf{+4.684 pp} & 43/44 & 40/44 & \textbf{40/44} \\
Gaussian-random & 4.780\% & $-0.225$ pp & \textbf{+4.909 pp} & 43/44 & 43/44 & \textbf{43/44} \\
Lowpass-random & 5.941\% & $+0.936$ pp & \textbf{+3.748 pp} & 37/44 & 35/44 & \textbf{35/44} \\
Spectrum-random & 7.570\% & $+2.565$ pp & \textbf{+2.119 pp} & 37/44 & 27/44 & \textbf{22/44} \\
Global mean & 5.207\% & $+0.202$ pp & \textbf{+4.482 pp} & 38/44 & -- & -- \\
Wrong target$^\dagger$ & 6.209\% & $+1.204$ pp & \textbf{+3.588 pp} & 32/40 & -- & -- \\
\bottomrule
\end{tabular}

\vspace{2pt}
\footnotesize
The Gaussian-random and spectrum-random tests have minimum attainable $p=1/101=.0099$; lowpass-random has minimum attainable $p=1/31=.0323$. $^\dagger$Four unavailable wrong-target comparisons are excluded.
\end{table*}

\begin{table}[t]
\centering
\small
\setlength{\tabcolsep}{2.6pt}
\caption{Target-specific FPR inflation by victim architecture. Each model is evaluated on the same 11 frozen candidates. Corrected counts are after Benjamini--Hochberg correction. RegNet-Y-8GF and PVT-v2-B2 are evaluated only after candidate selection and the analysis protocol are frozen.}
\label{tab:model-transfer}
\begin{tabular}{@{}lrrrr@{}}
\toprule
Model & Pert. & $\Delta$FPR & Positive & Corrected \\
      & FPR & & cells & $q\leq.05$ \\
\midrule
ResNet-50  & 5.912\%  & +0.907 pp  & 10/11 & 9/11 \\
ConvNeXt-T & 6.117\%  & +1.112 pp  & 11/11 & 9/11 \\
ViT-B/16   & 11.615\% & +6.610 pp  & 11/11 & 11/11 \\
Swin-T     & 15.111\% & +10.106 pp & 11/11 & 11/11 \\
\midrule
\multicolumn{5}{@{}l}{\textit{Prospectively held-out architectures}} \\
RegNet-Y-8GF & 7.731\% & +2.726 pp & 11/11 & 11/11 \\
PVT-v2-B2 & 15.035\% & +10.030 pp & 11/11 & 11/11 \\
\bottomrule
\end{tabular}
\end{table}

\section{Results}
\label{sec:results}
We first report whether raw class averages are recognizable to the models and which frequency components of the mean-based directions appear useful. We then report the confirmation-slice results and a selection-independent summary of the staged candidate panels on the frozen candidate panel. We follow this with comparisons on the matched controls in addition to a breakdown by construction, architecture, and secondary attack metrics, including two prospectively held-out architectures. Figure~\ref{fig:qualitative-results} shows representative confirmation-slice examples used to illustrate these aggregate effects.

Raw class-average images are rarely recognized as their own classes, indicating that the final effect requires the class contrasts, whitening, and frequency selection used by the retained constructions; full results are reported in Appendix Table~\ref{tab:mean}.

\subsection{Proposed Directions Increase Target-Specific FPR}
The frozen confirmation panel contains 11 target-construction-budget candidates evaluated on four models, for 44 model-candidate cells. Each cell is evaluated on 9,990 target-negative confirmation images, giving 439,560 condition-level image evaluations. These are condition-level evaluations rather than unique images, since the same validation image can appear under different targets, constructions, budgets, and models.

Our confirmation panel contains 22,000 targeted false-positives on our clean, unperturbed images. After applying the proposed directions, this increases to 42,589 target false-positive decisions. This is a net increase of 20,589 false positives, or 46.84 additional false positives per 1,000 target-negative images. In other words, the proposed directions increase target-specific FPR from 5.005\% to 9.689\%, a 1.94$\times$ lift over the clean baseline (Table~\ref{tab:headline-fpr}). The increase is positive in 43 of 44 cells, and 40 of 44 clean-to-perturbed comparisons remain significant after Benjamini--Hochberg correction.

\begin{figure}[t]
\centering
\includegraphics[width=\linewidth]{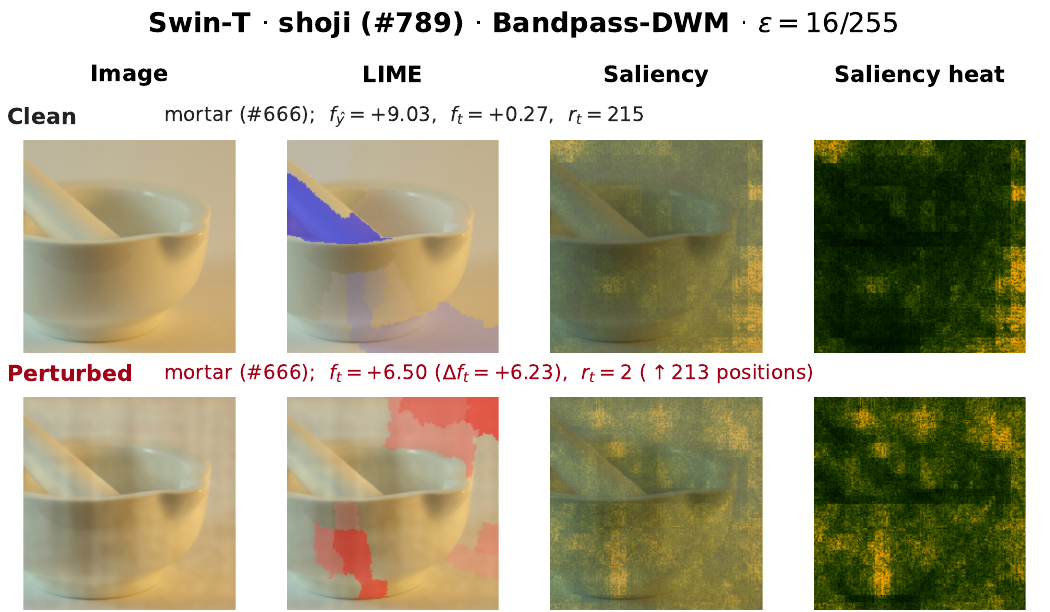}
\caption{One representative interpretability example for the frozen Swin-T shoji candidate using bandpass-DWM at $\epsilon=16/255$. The top-1 prediction remains mortar, while the target-logit lift is $+6.23$ and the target rank improves by 213 positions, from 215 to 2. Red LIME regions support the target class and blue regions oppose it; saliency views are unsigned.}
\label{fig:interpretability-example}
\end{figure}

The target-specific increase is not accompanied by a general increase in false positives for arbitrary wrong classes. The generic any-wrong-class FPR decreases from 5.005\% to 4.452\%. Figure~\ref{fig:interpretability-example} shows one representative target-directed attribution shift; the full three-example panel appears in Appendix Figure~\ref{fig:interpretability-results}.

\subsection{\texorpdfstring{Selection-Independent Candidate Summary}{Selection-Independent Candidate Summary}}

The staged search narrows 100 targets and 800 target--construction--budget configurations to 11 frozen candidates. Positive target-FPR movement occurs beyond the final panel, while effect size increases through successive narrowing; full funnel and distributional results are reported in Appendix Table~\ref{tab:selection-independent} and Figure~\ref{fig:selection-stage-ecdf}. Across all 64 concept-check configurations, pairwise Spearman correlations across the four development architectures range from 0.62 to 0.82 (Appendix Figure~\ref{fig:concept-check-correlation}). The broad screen is descriptive rather than an ImageNet-wide prevalence estimate.

\subsection{Controls Explain Some, but Not All, of the Effect}

The matched controls demonstrate several findings with our controls. First, Gaussian-random noise does not reproduce the effect. Its average perturbed FPR is 4.780\%, which is slightly below our clean baseline. Our proposed directions exceed the Gaussian-random mean in 43 of 44 cells. Because this control has 100 seeds, the smallest possible empirical $p$-value is $1/101=0.0099$. Forty-three of 44 comparisons are significant after Benjamini--Hochberg correction, and 42 of 44 also have positive proposed FPR movement.

Lowpass-random perturbations produce some false-positive inflation, raising FPR to 5.941\%. The proposed directions remain larger by 3.748 percentage points on average and exceed the lowpass-random mean in 37 of 44 cells. Thirty-five of those comparisons remain significant after Benjamini--Hochberg correction. While this shows that smooth low-frequency noise can move target scores, its shift is relatively negligible and it significantly underperforms our proposed method.

Spectrum-random is the strongest control. It raises FPR to 7.570\%, which means that matching the proposed direction's frequency magnitude accounts for part of the effect. Even so, the proposed directions remain larger on average by 2.119 percentage points and exceed the spectrum-random mean in 37 of 44 cells. Spectrum-random reproduces 54.8\% of the proposed mean FPR increase. Although 27 comparisons have raw $p\leq.05$, 22 survive Benjamini--Hochberg correction. Therefore, we treat spectrum matching as an important partial explanation while retaining a corrected residual in half of the original model--candidate cells.

A plausible interpretation is that the retained constructions align with class-associated spatial-frequency cues already used by the classifiers: FFT-Hellinger emphasizes target-distinctive frequency bands, while bandpass whitening emphasizes target-versus-complement deviations that are large relative to ordinary source-data variance. Spectrum-random preserves each candidate's Fourier magnitude and therefore much of the energy allocation that can exploit model spectral sensitivity, but phase randomization removes its target-conditioned spatial organization. The corrected residual in 22 of 44 cells is consistent with that organization contributing beyond magnitude alone, although this remains a mechanistic hypothesis rather than a causal decomposition.

Global-mean and wrong-target controls are also weaker than the proposed direction, supporting a target-specific rather than class-agnostic effect (Table~\ref{tab:control-comparison}); detailed interpretation is provided in Appendix~\ref{app:controls}.

Construction-specific confirmation counts are reported in Table~\ref{tab:headline-fpr}; detailed matched budget, construction, and exploratory frequency-ablation results are reported in Appendix Tables~\ref{tab:matched-construction} and~\ref{tab:frequency-ablation}.

\subsection{The Effect Transfers Across Architectures}

The same frozen perturbations are evaluated across the four development architectures and, without reselection, two prospectively held-out architectures. On held-out RegNet-Y-8GF and PVT-v2-B2, mean FPR rises by 2.726 and 10.030 percentage points, respectively; all 22 held-out model--candidate cells show positive movement, survive paired Benjamini--Hochberg correction, and exceed the 100-draw Gaussian null after correction. Effect magnitude remains architecture-dependent (Table~\ref{tab:model-transfer}).

The descriptive candidate-level association is also strong: mean target-FPR movement on the original four architectures correlates with mean movement on the held-out pair (Pearson $r=0.940$, Spearman $\rho=0.909$; Appendix Figure~\ref{fig:heldout-transfer-correlation}), indicating that candidate-strength ordering is largely preserved across the two panels.

%% file: sec/7_conclusion.tex
\section{Conclusion}
\label{sec:conclusion}

We show that source-statistical perturbations can induce target-specific false positives without model gradients, queries, or image-specific optimization. On the frozen confirmation panel, FFT-Hellinger and bandpass-whitened directions increase target-specific FPR from 5.005\% to 9.689\%, with positive movement in 43 of 44 model--candidate cells and 40 cells remaining significant after Benjamini--Hochberg correction. The same 11 candidates produce positive movement in all 22 evaluations on RegNet-Y-8GF and PVT-v2-B2 after both architectures are held out from candidate development. The effect is strongest for thresholded FPR, target logits, and target ranks.

The controls clarify what this result does and does not show. Gaussian random noise and global mean remain near the clean baseline, while lowpass-random produces a smaller effect. With 100 Gaussian draws, 42 of 44 original cells show both positive FPR movement and a corrected advantage, and all 22 held-out cells are positive and significant. Spectrum-random is the strongest control and explains part of the behavior, but the proposed directions remain larger on average and exceed the spectrum-random mean in most cells. With 100 spectrum draws, 22 of 44 original cells retain a Benjamini--Hochberg-significant residual, while spectrum matching reproduces 54.8\% of the proposed mean FPR increase. The current evidence therefore supports the existence of selected, transferable target-specific false-positive vulnerabilities induced by class-conditioned image statistics.

%% file: appendix.tex
\clearpage
\appendix

\section{\texorpdfstring{Prior-Work Comparison}{Prior-Work Comparison}}
\label{app:rq-prior}

Table~\ref{tab:ours-but-not-theirs} summarizes the closest prior-work families and how our setting differs from them.

\input{tbl/core_contributions_full.tex}

\section{\texorpdfstring{Detailed Problem Formulation}{Detailed Problem Formulation}}
\label{app:problem-details}

\subsection{\texorpdfstring{Target rank and secondary metrics}{Target rank and secondary metrics}}
\label{app:secondary-metrics}

We summarize the standing of a target class $t$ on a \emph{single image} by the integer $r^{(m)}_t(\boldsymbol{x})$, defined as the position of $t$ in the descending ordering of the logits $f^{(m)}_1(\boldsymbol{x}),\dots,f^{(m)}_K(\boldsymbol{x})$,
\begin{equation}
  r^{(m)}_t(\boldsymbol{x})
  = 1 + \sum_{k=1}^{K}
      \mathbbm{1}\bigl\{f^{(m)}_k(\boldsymbol{x}) > f^{(m)}_t(\boldsymbol{x})\bigr\}.
  \label{eq:r}
\end{equation}
Here $r^{(m)}_t(\boldsymbol{x})\in\{1,\dots,K\}$: it equals $1$ when $t$ is the top-scoring class and $K$ when $t$ is scored last, so a smaller $r$ means a stronger standing for $t$. A perturbation that promotes $t$ \emph{decreases} $r$; the amount by which it decreases is the \emph{$r$-improvement} $\Delta r$.

For a model $f^{(m)}$ and image $\boldsymbol{x}\in\mathcal{V}_t$ with perturbed counterpart $\tilde{\boldsymbol{x}}$ we record three per-image quantities, each averaged over $\mathcal{V}_t$. The \emph{target-logit shift}
\begin{equation}
  \Delta z^{(m)}_t(\boldsymbol{x})
    = f^{(m)}_t(\tilde{\boldsymbol{x}})-f^{(m)}_t(\boldsymbol{x})
\end{equation}
is positive when the perturbation raises the score of $t$. The \emph{$r$-improvement}
\begin{equation}
  \Delta r^{(m)}_t(\boldsymbol{x})
    = r^{(m)}_t(\boldsymbol{x})-r^{(m)}_t(\tilde{\boldsymbol{x}})
\end{equation}
counts how many positions class $t$ climbs in the logit ordering, and is positive when $t$ is promoted. The \emph{top-$k$ targeted success}
\begin{equation}
  \mathrm{ASR}^{(m)}_k
    = \mathbb{E}_{\mathcal{V}_t}\,
      \mathbbm{1}\bigl\{r^{(m)}_t(\tilde{\boldsymbol{x}})\le k\bigr\},
  \qquad k\in\{1,5\},
\end{equation}
is the fraction of target-negative images for which $t$ is pushed into the top $k$ predictions.

\subsection{\texorpdfstring{Calibrated FPR threshold}{Calibrated FPR threshold}}
\label{app:fpr-threshold}

On the clean target-negative images we fix a threshold $\tau^{(m)}_t$ at the $(1-\alpha)$ empirical quantile of the clean target logits, with $\alpha=0.05$,
\begin{equation}
  \tau^{(m)}_t:\quad
  \mathbb{E}_{\mathcal{V}_t}\,
    \mathbbm{1}\bigl\{f^{(m)}_t(\boldsymbol{x})>\tau^{(m)}_t\bigr\}=\alpha,
\end{equation}
so the clean false-positive rate equals $\alpha$ by construction. We report $\Delta\mathrm{FPR}_\alpha$ both in percentage points and as $1000\,\Delta\mathrm{FPR}_\alpha$ extra false fires per $1000$ images, with a paired nonparametric bootstrap confidence interval over $\mathcal{V}_t$. To confirm that the effect is genuinely targeted, we verify that the untargeted any-wrong-class false-positive rate is not inflated.

\subsection{\texorpdfstring{Controls}{Controls}}
\label{app:controls}


We compare each $\boldsymbol{\delta}_t$ against budget-matched controls that share its $\ell_\infty$ norm but lack the target-class signal. All controls are projected to the same budget $\varepsilon$ via \eqref{eq:proj}, and the spectral and statistical operators they invoke are defined in \cref{app:operator-details}.

\begin{itemize}[leftmargin=1.4em,itemsep=2pt,topsep=2pt]
  \item \textbf{random}: $\boldsymbol{v}\sim\mathrm{Unif}[-1,1]^{d}$, an unstructured baseline.
  \item \textbf{lowpass-random}: $G_{\mathrm{low}}$ applied to a random direction, matching the smoothness of low-frequency perturbations without any class signal.
  \item \textbf{spectrum-random}: a phase-randomized version of $\boldsymbol{\delta}_t$ that preserves its power spectrum, $\mathcal{F}(\boldsymbol{v})=\lvert\mathcal{F}(\boldsymbol{\delta}_t)\rvert\odot e^{i\theta}$ with $\theta\sim\mathrm{Unif}[0,2\pi)$ under Hermitian symmetry, isolating spectral magnitude from spatial structure.
  \item \textbf{global / statistical}: the dataset-wide mean image $\boldsymbol{v}=\boldsymbol{\mu}$, a class-agnostic statistical direction.
  \item \textbf{wrong-target}: a perturbation $\boldsymbol{\delta}_{t'}$ built for a different class $t'\neq t$, testing target specificity.
\end{itemize}

The deterministic controls are much weaker. The global-mean control barely outperforms the clean baseline, raising FPR only to 5.207\%. The wrong-target control raises FPR to 6.209\% over the 40 valid cells, but the proposed direction still exceeds it in 32 of those 40 cells. This indicates that target identity matters. Simply applying another class-statistical direction is not equivalent to applying the direction for the intended target. Table~\ref{tab:control-comparison} summarizes these matched-control comparisons.

\subsection{\texorpdfstring{Selection and inference protocol}{Selection and inference protocol}}
\label{app:protocol}

For any metric $\Theta$ we draw $S$ seeds per control to form a null distribution $\{\Theta^{\mathrm{null}}_s\}_{s=1}^{S}$, and summarize the position of the observed value $\Theta^{\mathrm{main}}$ by
\begin{equation}
  \hat{p}
    = \frac{1+\lvert\{s:\Theta^{\mathrm{null}}_s\ge\Theta^{\mathrm{main}}\}\rvert}{S+1},
  \qquad
  z = \frac{\Theta^{\mathrm{main}}-\bar{\Theta}^{\mathrm{null}}}{\operatorname{sd}(\Theta^{\mathrm{null}})}.
\end{equation}
The $p$-value is floored at $1/(S+1)$, so $z$ quantifies the margin by which the perturbation exceeds the null. Although construction is model-free, the reported triple $(t,\text{method},\varepsilon)$ is selected from a candidate grid by a model-dependent score; to guard against this selection we report a candidate as \emph{transferring} only when its paired $\Delta\mathrm{FPR}_\alpha$ confidence interval excludes $0$ \emph{and} it exceeds every null seed, simultaneously on all $M$ models. For the revised held-out experiment, the 11-candidate panel and analysis protocol are frozen before RegNet-Y-8GF and PVT-v2-B2 are evaluated, and paired and Gaussian tests are corrected separately across the resulting 22 held-out cells.

\section{\texorpdfstring{Detailed Construction Operators}{Detailed Construction Operators}}
\label{app:operator-details}

Each operator takes an image-shaped direction $\boldsymbol{v}\in\mathbb{R}^{C\times H\times W}$, where $\boldsymbol{v}$ is either the contrast $\boldsymbol{g}_t$ or its whitened form $\boldsymbol{w}_t$, and retains a chosen part of it.

\paragraph{Spectral operators.}
Let $\mathcal{F}$ denote the two-dimensional discrete Fourier transform applied independently to each color channel, and $\mathcal{F}^{-1}$ its inverse; thus $\mathcal{F}$ maps a channel from pixel space into frequency space and $\mathcal{F}^{-1}$ maps it back. For a frequency location $\boldsymbol{\xi}$ we write its radial frequency $\rho(\boldsymbol{\xi})=\lVert\boldsymbol{\xi}\rVert_2$ (the distance from the zero, or DC, frequency), and define the radial band mask
\begin{equation}
  \mathcal{M}_{[a,b]}(\boldsymbol{\xi})
    = \mathbbm{1}\bigl\{a\le\rho(\boldsymbol{\xi})\le b\bigr\}.
\end{equation}
Multiplying a spectrum by such a mask and inverting yields a band-limited operator; the low-pass, high-pass, and band-pass operators are
\begin{align}
  G_{\mathrm{low}}(\boldsymbol{v})
    &= \mathcal{F}^{-1}\!\bigl(\mathcal{M}_{[0,\rho_c]}\odot\mathcal{F}\boldsymbol{v}\bigr),\\
  G_{\mathrm{high}}(\boldsymbol{v})
    &= \mathcal{F}^{-1}\!\bigl(\mathcal{M}_{[\rho_c,\infty)}\odot\mathcal{F}\boldsymbol{v}\bigr),\\
  G_{\mathrm{band}}(\boldsymbol{v})
    &= \mathcal{F}^{-1}\!\bigl(\mathcal{M}_{[\rho_\ell,\rho_h]}\odot\mathcal{F}\boldsymbol{v}\bigr),
\end{align}
with cutoff $\rho_c$ and band edges $\rho_\ell<\rho_h$. Intuitively, $G_{\mathrm{low}}$ keeps smooth, blurry structure, $G_{\mathrm{high}}$ keeps sharp edges and texture, and $G_{\mathrm{band}}$ keeps mid-frequency structure.

\paragraph{Channel and DC operators.}
The per-channel mean (DC component) of a direction is
\begin{equation}
  \mathrm{DC}(\boldsymbol{v})_c
    = \frac{1}{HW}\sum_{h,w} v_{c,h,w}.
\end{equation}
From it we form the \emph{zero-DC} operator, which removes each channel's average and keeps only spatial pattern,
\begin{equation}
  \mathrm{ZDC}(\boldsymbol{v})_c
    = \boldsymbol{v}_c-\mathrm{DC}(\boldsymbol{v})_c\,\boldsymbol{1},
\end{equation}
and the \emph{channel-mean} operator, which keeps only the overall per-channel color shift,
\begin{equation}
  \mathrm{CM}(\boldsymbol{v})_c
    = \mathrm{DC}(\boldsymbol{v})_c\,\boldsymbol{1},
\end{equation}
where $\boldsymbol{1}$ is the constant all-ones image. Thus $\mathrm{ZDC}$ discards a global color cast while preserving structure, and $\mathrm{CM}$ does the opposite.

\paragraph{FFT--Hellinger operator.}
This operator uses $\mathcal{D}_{\mathrm{tr}}$ to identify the \emph{radial frequency bands} whose share of spectral power differs most between target-class images and the dataset as a whole, and reweights the contrast $\boldsymbol{g}_t$ to emphasize those bands. Reusing the Fourier transform $\mathcal{F}$ and radial frequency $\rho(\boldsymbol{\xi})$ above, we partition the frequency plane into $B$ radial bands and write $b(\boldsymbol{\xi})\in\{1,\dots,B\}$ for the band containing $\boldsymbol{\xi}$. The radial power of an image $\boldsymbol{x}$ in band $b$ is its channel-averaged Fourier power summed over that band,
\begin{equation}
  P_b(\boldsymbol{x})
    = \sum_{\boldsymbol{\xi}:\,b(\boldsymbol{\xi})=b}
      \frac{1}{C}\sum_{c=1}^{C}
      \bigl\lvert(\mathcal{F}\boldsymbol{x})_{c}(\boldsymbol{\xi})\bigr\rvert^{2}.
\end{equation}
Averaging $P_b$ over the target-class images and over all images yields the two radial power profiles $\bar{P}^{t}=(\bar{P}^{t}_b)_b$ and $\bar{P}=(\bar{P}_b)_b$, which we normalize into probability vectors over the $B$ bands,
\begin{equation}
  p^{t}_b=\frac{\bar{P}^{t}_b}{\sum_{b'}\bar{P}^{t}_{b'}},
  \qquad
  p_b=\frac{\bar{P}_b}{\sum_{b'}\bar{P}_{b'}}.
\end{equation}
We measure the discriminativeness of each band through the Hellinger chart, i.e. the signed difference of the square-root profiles, mean-centered so the weights add no net energy,
\begin{equation}
  h_t(b)
    = \Bigl(\sqrt{p^{t}_b}-\sqrt{p_b}\Bigr)
      - \frac{1}{B}\sum_{b'=1}^{B}\Bigl(\sqrt{p^{t}_{b'}}-\sqrt{p_{b'}}\Bigr).
\end{equation}
A positive $h_t(b)$ marks a band carrying relatively more target power than the dataset average and a negative one a band carrying relatively less; before centering, $\tfrac{1}{2}\sum_b\bigl(\sqrt{p^{t}_b}-\sqrt{p_b}\bigr)^2$ is exactly the squared Hellinger distance between the two profiles. Broadcasting the band weights back to the frequency plane as $W(\boldsymbol{\xi})=h_t\bigl(b(\boldsymbol{\xi})\bigr)$, the operator modulates the spectrum of the contrast, inverts, and removes each channel's spatial mean,
\begin{equation}
  \boldsymbol{v}^{\mathrm{fft\text{-}hel}}_t
    = \mathrm{ZDC}\!\Bigl(
        \mathcal{F}^{-1}\!\bigl(W\odot\mathcal{F}\boldsymbol{g}_t\bigr)
      \Bigr).
\end{equation}

Table~\ref{tab:families} summarizes the perturbation families used in the candidate screen and confirmation panel.

\begin{table}[t]
  \centering
  \begin{tabular}{@{}ll@{}}
    \toprule
    Method & $\boldsymbol{v}_t$\\
    \midrule
    \texttt{mean} & $\boldsymbol{g}_t$\\
    \texttt{channel-mean} & $\mathrm{CM}(\boldsymbol{g}_t)$\\
    \texttt{zero-dc-mean} & $\mathrm{ZDC}(\boldsymbol{g}_t)$\\
    \texttt{lowpass-mean} & $G_{\mathrm{low}}(\boldsymbol{g}_t)$\\
    \texttt{highpass-mean} & $G_{\mathrm{high}}(\boldsymbol{g}_t)$\\
    \texttt{bandpass-mean} & $G_{\mathrm{band}}(\boldsymbol{g}_t)$\\
    \texttt{sign-mean} & $\operatorname{sign}(\boldsymbol{g}_t)$\\
    \texttt{fft-hellinger} & $\boldsymbol{v}^{\mathrm{fft\text{-}hel}}_t$\\
    \texttt{diag-whitened-mean} & $\boldsymbol{w}_t$\\
    \texttt{bandpass-diag-whitened-mean} & $G_{\mathrm{band}}(\boldsymbol{w}_t)$\\
    \bottomrule
  \end{tabular}
  \caption{Perturbation families as operators on the contrast $\boldsymbol{g}_t$ or its diagonally whitened form $\boldsymbol{w}_t$.}
  \label{tab:families}
\end{table}

\section{\texorpdfstring{Detailed Experimental Protocol}{Detailed Experimental Protocol}}
\label{app:protocol-details}

\paragraph{Validation slices.}
Thus, we have to separate the datasets we use for candidate development and final evaluation. In order to achieve this in a random but still reproducible way, we shuffle the 50 validation images within each class using seed 17 and divide the resulting ordering into non-overlapping slices.

Positions 0--4 form the \emph{concept-check slice}. They contain five images per class and 5,000 images in total. We use this slice to evaluate an initial set of candidates, check whether they produce target-directed changes in FPR, logits, and ranks. We also compare them with matched controls, and remove candidates that are ineffective or affected by obvious implementation artifacts.

Positions 5--14 form the \emph{candidate-validation slice}. They contain 10 images per class and 10,000 images in total. We use this larger, disjoint slice to reevaluate the retained candidates with additional random-control seeds and to select the final target--construction--budget combinations.

Positions 15--24 form the \emph{confirmation slice}, also containing 10 images per class and 10,000 images in total. The candidate panel is fixed before this slice is evaluated, and only results from this slice are used for the headline confirmation analysis.

Positions 25--49 comprise the remaining 25,000 validation images. They are not used in the current experiments.

\paragraph{Class-average prototype check.}
Because several of our perturbation families are derived from class-level image averages, we first evaluate whether raw class averages contain any model-visible evidence of their own class. We construct one average image for each of the 1,000 ImageNet classes using all available training images from that class. We then evaluate each class-average image on all four victim models, for a total of 4,000 model--prototype evaluations and record whether the corresponding class appears in the model's top-1 or top-5 predictions along with its logit and rank. Table~\ref{tab:mean} reports this class-average recognition check.

\begin{table}[t]
\centering
\small
\setlength{\tabcolsep}{3.2pt}
\caption{Recognition of ImageNet class-average images. Each model is evaluated on the mean image of each of the 1,000 ImageNet classes.}
\label{tab:mean}
\begin{tabular}{@{}lrrrr@{}}
\toprule
Model & Top-1 & Top-5 & Mean & Median \\
      & (\%)  & (\%)  & rank & rank \\
\midrule
ResNet-50  & 1.3 & 4.2 & 329.8 & 255.5 \\
ConvNeXt-T & 1.8 & 5.7 & 334.3 & 282.0 \\
ViT-B/16   & 1.1 & 4.2 & 343.5 & 251.0 \\
Swin-T     & 2.5 & 5.6 & 293.0 & 212.5 \\
\bottomrule
\end{tabular}
\end{table}

The raw class-average images are rarely classified as their own class. Across the four victim models, own-class top-1 accuracy ranges from 1.1\% to 2.5\%, and own-class top-5 accuracy ranges from 4.2\% to 5.7\%. The median own-class rank is also far from the top of the prediction list, ranging from 212.5 to 282.0. This suggests that the final effect is not explained by class averages simply looking like recognizable examples of their corresponding classes. Instead, the useful signal appears to require the transformations described in Section~\ref{sec:construction}, such as class contrasts, whitening, and frequency selection.

\paragraph{Exploratory frequency ablation.}
Alongside the main runs, we ran a small screen on the mean-based perturbation. The four versions differed only in the frequency filter applied before normalization: no filtering for the raw mean, removal of the image-wide offset for zero-DC, retention of lower-to-middle spatial frequencies for bandpass, and removal of frequencies below the highpass cutoff for highpass. We used the screen to choose which mean-based variants to carry into the confirmation run.

This differs from the lowpass-random control used in the confirmation experiment. The frequency ablation still starts from the class-statistical direction; the random control starts from noise. Thus, the ablation asks which part of our direction matters, while the random control asks whether a non-class-statistical perturbation with similar smoothness can produce the same effect.

This experiment uses three target classes, ResNet-50 and ViT-B/16, and perturbation budgets of $8/255$, $16/255$, and $32/255$. The resulting grid contains $3\times2\times4\times3=72$ proposed-direction conditions. Each condition is evaluated on 1,998 target-negative validation images. We use this stage to identify which frequency band is carried forward into the confirmation experiment. Table~\ref{tab:frequency-ablation} reports the resulting exploratory frequency ablation.

\begin{table}[t]
\centering
\small
\setlength{\tabcolsep}{4pt}
\caption{Exploratory frequency ablation. Values are means over 18 model-target-budget conditions per construction. FPR changes are percentage points.}
\label{tab:frequency-ablation}
\begin{tabular}{@{}lrrr@{}}
\toprule
Construction & $\Delta$FPR@5\% & $\Delta$logit & $\Delta$rank \\
\midrule
Bandpass mean & +0.943 & +0.028 & +14.6 \\
Raw mean      & +0.731 & +0.026 & +16.4 \\
Zero-DC mean  & +0.759 & +0.017 & +8.7 \\
Highpass mean & -0.206 & -0.012 & -8.4 \\
\bottomrule
\end{tabular}
\end{table}

The exploratory frequency ablation shows that the effect is not driven by high-frequency residuals. Bandpass, raw mean, and zero-DC mean directions all increase target-specific FPR on average, while the highpass mean direction decreases it. Highpass also produces negative average target-logit and target-rank movement. We therefore carry forward the frequency-restricted bandpass direction, along with the FFT-Hellinger construction, into the confirmation experiment. We use this ablation to motivate method selection. It is exploratory and not used as evidence for the final claims.

\begin{table*}[t]
\centering
\small
\caption{Frozen target--construction panel evaluated on the confirmation slice. Each candidate is evaluated on four development models; the same panel is subsequently evaluated on two prospectively held-out architectures.}
\label{tab:confirmation-panel}
\begin{tabular}{llp{7.0cm}cc}
\toprule
Construction & $\epsilon$ & Targets & Candidates & Model cells \\
\midrule
FFT-Hellinger & $8/255$ & 267 (standard poodle), 789 (shoji) & 2 & 8 \\
FFT-Hellinger & $16/255$ & 140 (red-backed sandpiper), 207 (golden retriever), 267 (standard poodle), 789 (shoji) & 4 & 16 \\
Bandpass diagonal-whitened mean & $8/255$ & 789 (shoji) & 1 & 4 \\
Bandpass diagonal-whitened mean & $16/255$ & 140 (red-backed sandpiper), 207 (golden retriever), 384 (indri), 789 (shoji) & 4 & 16 \\
\midrule
Total & & & 11 & 44 \\
\bottomrule
\end{tabular}
\end{table*}

\paragraph{Broad candidate screen.}
We conduct an initial screen over 100 target classes, consisting of five previously identified prototype candidates and 95 targets selected using a fixed random sample. For each target, we evaluate zero-DC mean, diagonal-whitened mean, bandpass diagonal-whitened mean, and FFT-Hellinger directions at $8/255$ and $16/255$. The screen uses ResNet-50, ViT-B/16, and one validation image per class. For every target, perturbation method, perturbation budget, and victim model, we generate one lowpass-random direction and one spectrum-random direction. This stage is used to reduce a broad target and method search to a smaller set of candidates. The complete broad panel contains 800 target--construction--budget configurations and 1,600 model--candidate cells; eight targets are selected, and all 64 construction--budget configurations associated with those targets are carried into concept check.

\paragraph{Focused concept check.}
Eight targets from the broad screen are subsequently used in the concept-check slice described above. We evaluate four candidate constructions, two perturbation budgets, and all four victim models, giving $8\times4\times2\times4=256$ model-level proposed-direction conditions. Stochastic controls are evaluated with 10 seeds in this stage. We retain candidates that show positive target-FPR, target-logit, and target-rank movement, compare favorably with the lowpass-random and spectrum-random controls, transfer across more than one model family, and do not exhibit obvious implementation artifacts or severe corruption. Candidate selection at this stage is rule-guided rather than based on a preregistered scalar score. This stage narrows the panel to four targets, two constructions, and two budgets, producing 16 candidate-validation configurations.

\paragraph{Candidate validation.}
The retained candidates are reevaluated on the larger candidate-validation slice using all four victim models. This stage focuses on FFT-Hellinger and bandpass diagonal-whitened mean at $8/255$ and $16/255$. Gaussian-random controls use 10 seeds per condition, while lowpass-random and spectrum-random controls use 30 seeds each. Global-mean and wrong-target controls are deterministic. Results from this stage are used to freeze the final target--construction--budget panel before the confirmation slice is evaluated. Nine candidate-validation configurations are retained, and two target-207 continuity configurations are reintroduced from concept check, yielding 11 frozen candidates. Table~\ref{tab:confirmation-panel} lists the frozen panel used for the confirmation-slice evaluation.

\paragraph{Confirmation controls.}
Each proposed direction is compared with controls evaluated on the same target class, target-negative images, victim model and perturbation budget. We use 100 independently randomly sampled pixel-space noise directions for the \emph{Gaussian-random control}. The \emph{lowpass-random} control uses 30 random directions smoothed towards low spatial frequencies to test if smooth random structure is sufficient. The \emph{spectrum-random} control uses 100 phase-randomized directions with the same Fourier magnitude as the proposed direction, testing whether only matching the frequency content can explain the direction. The 100-draw Gaussian control is evaluated on all six models; lowpass-random and spectrum-random remain restricted to the four development models. The global-mean control uses ImageNet-wide statistics rather than target-specific statistics. The wrong-target control uses a source-statistical direction from a different target class.

There is no wrong target comparison for the bandpass-whitened $8/255$ target-789 candidate because it is the only target in that construction--budget group. Thus, we omit the wrong-target control for those four model cells but retain its proposed direction and all of its other corresponding controls.

\paragraph{Statistical tests and Benjamini--Hochberg correction.}
For each model--candidate pair, we compare how often an image crosses the target threshold after perturbation with how often it falls back below the threshold. A newly crossed example is below the target threshold when clean and above it after perturbation. A recovered example is above the threshold when clean and below it after perturbation. Under the null hypothesis that the perturbation has no directional effect on target-threshold crossings, these two outcomes are equally likely. We therefore use a one-sided exact binomial test, equivalently a one-sided exact McNemar test, and apply Benjamini--Hochberg correction across the 44 confirmation cells so that the expected fraction of false discoveries among the significant cells is controlled at 5\%~\cite{BenjaminiHochberg1995}. The same paired test is applied to the 22 held-out model--candidate cells, with Benjamini--Hochberg correction performed separately for that prespecified family.

Because the panel comprises $m=44$ model--candidate cells, some form of multiplicity adjustment is required. We adopt the Benjamini--Hochberg procedure, which controls the false discovery rate, in preference to family-wise schemes such as Bonferroni. The latter sacrifices considerable power across a panel of this size, while the former limits the expected fraction of erroneous rejections among the cells declared significant, which is the quantity of interest for a confirmatory panel. Ordering the one-sided $p$-values as $p_{(1)}\leq\cdots\leq p_{(m)}$, the procedure rejects every hypothesis up to the largest rank $k$ satisfying
\begin{equation}
p_{(k)} \leq \frac{k}{m}q, \qquad q=0.05.
\end{equation}
Tables~\ref{tab:headline-fpr}, \ref{tab:control-comparison} in the main paper report the corresponding Benjamini--Hochberg-corrected counts and control summaries. Table~\ref{tab:matched-construction} reports the matched budget and construction comparisons below.

For stochastic controls, we compare the proposed direction with the control-seed distribution within the same cell. The empirical one-sided $p$-value counts the fraction of control seeds whose FPR change is at least as large as the proposed direction, with a standard plus-one correction. The number of control seeds also limits the smallest empirical $p$-value we can report. With 100 Gaussian-random seeds, the smallest value is $1/101$. With 30 lowpass-random seeds, the smallest value is $1/31$; with 100 spectrum-random seeds, it is $1/101$. We apply Benjamini--Hochberg correction separately within the Gaussian-random, lowpass-random, and spectrum-random families, and separately for the original 44-cell and held-out 22-cell Gaussian families. Because global-mean and wrong-target controls are deterministic, we report paired effect differences for those controls instead of seed-based $p$-values.

\FloatBarrier
\section{\texorpdfstring{Revision Result Details}{Revision Result Details}}
\label{app:revision-results}

\subsection{\texorpdfstring{Selection-independent candidate summary}{Selection-independent candidate summary}}

\begin{figure}[t]
\centering
\includegraphics[width=\linewidth]{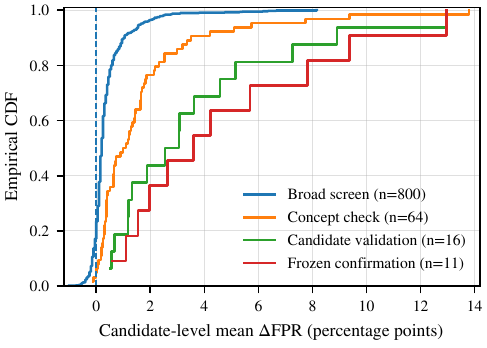}
\caption{Candidate-level empirical distributions of mean target-FPR movement at each stage. Curves move toward larger effects as the search narrows. The broad screen uses two models and a shallower evaluation than later stages, so stage differences combine selection with evaluation-depth differences.}
\label{fig:selection-stage-ecdf}
\end{figure}

To separate existence from prevalence, we summarize every evaluated configuration in each staged panel before subsequent narrowing. Positive target-FPR movement occurs beyond the final 11 candidates, while the mean and median effects increase as the search is narrowed (Table~\ref{tab:selection-independent} and Figure~\ref{fig:selection-stage-ecdf}). The broad screen is not an ImageNet-wide prevalence estimate because it uses two models, one validation image per class, and a target set containing five prototype targets plus 95 seeded-random targets.

\begin{table*}[t]
\centering
\small
\setlength{\tabcolsep}{3.0pt}
{
\caption{Candidate funnel and selection-independent target-specific FPR movement. Each row summarizes the complete evaluated panel before the next-stage decision. FPR changes are percentage points.}
\label{tab:selection-independent}
\begin{tabular}{@{}lrrrrrrp{0.22\textwidth}@{}}
\toprule
Stage & Targets & Configs. & Cells & Positive cells & Mean & Median & Next-stage decision \\
\midrule
Broad screen & 100 & 800 & 1,600 & 1,038/1,600 & +0.389 & +0.200 & Select 8 targets; carry their 64 configurations \\
Concept check & 8 & 64 & 256 & 239/256 & +1.688 & +0.521 & Retain 4 targets, 2 constructions, and 2 budgets (16 configurations) \\
Candidate validation & 4 & 16 & 64 & 63/64 & +3.654 & +1.281 & Retain 9 configurations; reintroduce 2 continuity configurations \\
Frozen confirmation & 5 & 11 & 44 & 43/44 & +4.684 & +1.567 & No further candidate selection \\
\bottomrule
\end{tabular}

\vspace{2pt}
\footnotesize
``Configs.'' denotes target--construction--budget configurations. Historical narrowing used multiple quantitative and visual criteria rather than one scalar threshold; two target-207 continuity configurations were reintroduced from concept check.}
\end{table*}

\subsection{\texorpdfstring{Qualitative interpretability example}{Qualitative interpretability example}}

\begin{figure*}[t]
\centering
\includegraphics[width=\textwidth]{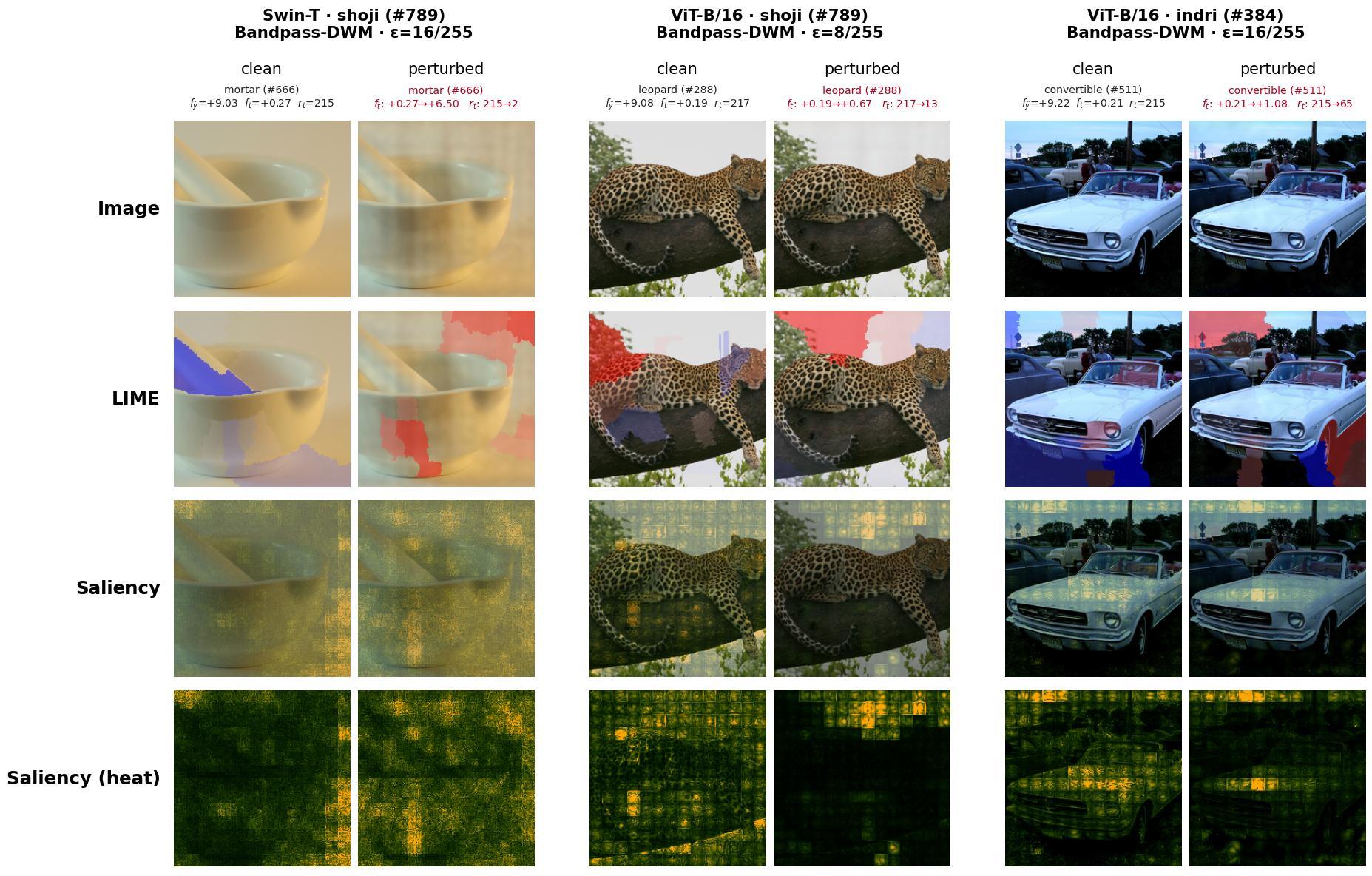}
\caption{Representative interpretability views for clean and perturbed inputs. Each column pair shows the clean and perturbed image for the same target, construction, budget, and victim model. Rows show the image, LIME overlay, saliency overlay, and saliency heat map. In the LIME row, red marks regions that support the target class and blue marks regions against it; the saliency rows are unsigned, so brighter yellow means larger attribution magnitude, regardless of direction. These visualizations are qualitative; aggregate confirmation statistics are reported in Tables~\ref{tab:headline-fpr}--\ref{tab:model-transfer}.}
\label{fig:interpretability-results}
\end{figure*}

Figure~\ref{fig:interpretability-results} provides a qualitative view of one such target-directed change.

\subsection{\texorpdfstring{FFT-Hellinger and Bandpass-Whitened Directions Both Contribute}{FFT-Hellinger and Bandpass-Whitened Directions Both Contribute}}

FFT-Hellinger and bandpass-whitened directions play different roles in the confirmation panel. FFT-Hellinger is the frequency-profile construction motivated by the class-level frequency-energy summaries. It is positive in all 24 of its cells, increases FPR from 5.005\% to 7.994\%, and is Benjamini--Hochberg significant against clean in 21 of 24 cells. Bandpass-whitened is the stronger empirical construction in this panel. It increases FPR from 5.005\% to 11.723\%, is positive in 19 of 20 cells, and is Benjamini--Hochberg significant against clean in 19 of 20 cells.

The confirmation panel is not balanced across methods, targets, and budgets. Thus, direct method comparisons should use matched subsets. For FFT-Hellinger, targets 267 and 789 appear at both $8/255$ and $16/255$. Within these matched cells, the $16/255$ result is larger for 7 out of 8 of them. For bandpass-whitened, target 789 appears at both budgets. The $16/255$ result is larger for all four models. At $16/255$, FFT-Hellinger and bandpass-whitened share targets 140, 207, and 789. On those matched cells, bandpass-whitened is larger for 10 out of 12 comparisons. These comparisons are descriptive because the candidate panel was selected during development. Table~\ref{tab:matched-construction} summarizes these matched subsets.

\begin{table*}[t]
\centering
\small
\setlength{\tabcolsep}{4pt}
\caption{Matched budget and construction comparisons. Values are mean target-specific FPR changes in percentage points.}
\label{tab:matched-construction}
\begin{tabular}{@{}p{0.22\textwidth}p{0.31\textwidth}p{0.16\textwidth}p{0.16\textwidth}r@{}}
\toprule
Comparison & Matched subset & Setting A & Setting B & B $>$ A \\
\midrule
FFT-Hellinger budget & Targets 267, 789; four models & $8/255$: 1.071 & $16/255$: 5.231 & 7/8 \\
Bandpass budget & Target 789; four models & $8/255$: 3.606 & $16/255$: 9.377 & 4/4 \\
Construction at $16/255$ & Targets 140, 207, 789; four models & FFT-Hell.: 4.385 & BP-whitened: 5.674 & 10/12 \\
\bottomrule
\end{tabular}
\end{table*}

\subsection{\texorpdfstring{Cross-model structure before final narrowing}{Cross-model structure before final narrowing}}

\begin{figure}[t]
\centering
\includegraphics[width=0.96\linewidth]{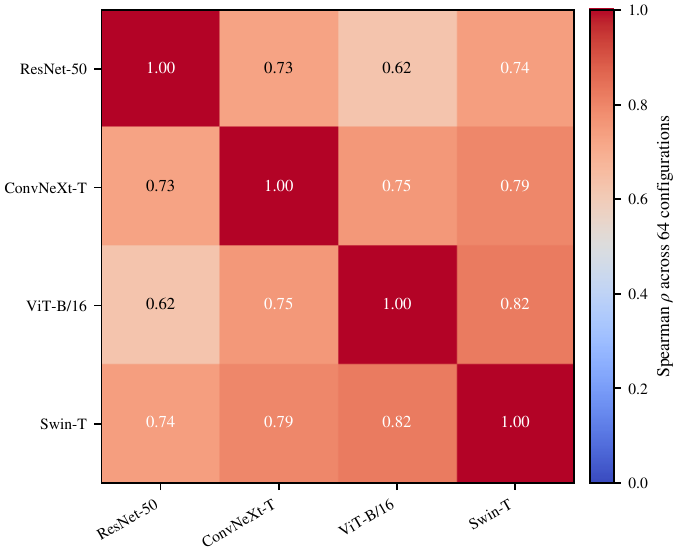}
\caption{Pairwise Spearman association of target-FPR movement across all 64 concept-check configurations, before narrowing to candidate validation.}
\label{fig:concept-check-correlation}
\end{figure}

Across all 64 concept-check configurations, model-specific target-FPR movements remain positively associated before narrowing to the 16 candidate-validation configurations. Pairwise Spearman correlations range from approximately 0.62 to 0.82 (Figure~\ref{fig:concept-check-correlation}), indicating that cross-model ordering is not created only by selecting the final 11 candidates. The eight target classes in this panel have already been selected during broad screening, so this is pre-final-selection rather than fully unselected evidence.

\subsection{\texorpdfstring{Held-out architecture transfer}{Held-out architecture transfer}}

The effect is positive for nearly every architecture-candidate pair, but its magnitude differs sharply by model. ResNet-50 and ConvNeXt-Tiny show smaller average increases, while ViT-B/16 and Swin-T are much more sensitive. On Swin-T, the mean target-specific FPR rises to 15.111\%, a 10.106 percentage-point increase over clean. On ViT-B/16, it rises to 11.615\%. The two convolutional models also show positive movement, but at a smaller scale (Table~\ref{tab:model-transfer}).

\begin{figure}[t]
\centering
\includegraphics[width=\linewidth]{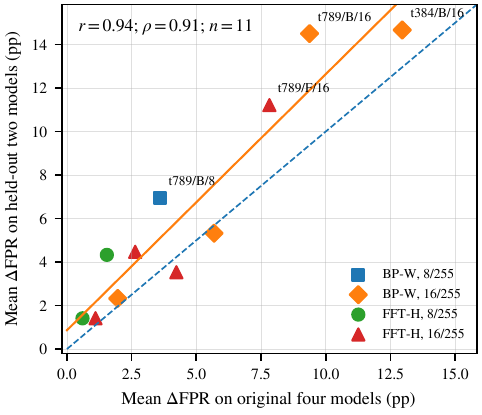}
\caption{Candidate-level association between mean target-FPR movement on the original four architectures and the two prospectively held-out architectures. Marker shape identifies construction and budget.}
\label{fig:heldout-transfer-correlation}
\end{figure}

Candidate strength on the original four architectures predicts candidate strength on the two prospectively held-out architectures (Figure~\ref{fig:heldout-transfer-correlation}). These correlations are descriptive because the panel contains 11 selected candidates.

\begin{table}[t]
\centering
\small
{
\caption{Descriptive association between candidate strength on the original and held-out architecture panels.}
\label{tab:heldout-correlation}
\begin{tabular}{@{}lrr@{}}
\toprule
Endpoint & Pearson $r$ & Spearman $\rho$ \\
\midrule
Target FPR movement & 0.940 & 0.909 \\
Target-logit shift & 0.933 & 0.845 \\
Target-rank improvement & 0.899 & 0.845 \\
\bottomrule
\end{tabular}}
\end{table}

\subsection{\texorpdfstring{Residual beyond spectrum matching}{Residual beyond spectrum matching}}

\begin{figure*}[t]
\centering
\includegraphics[width=0.96\textwidth]{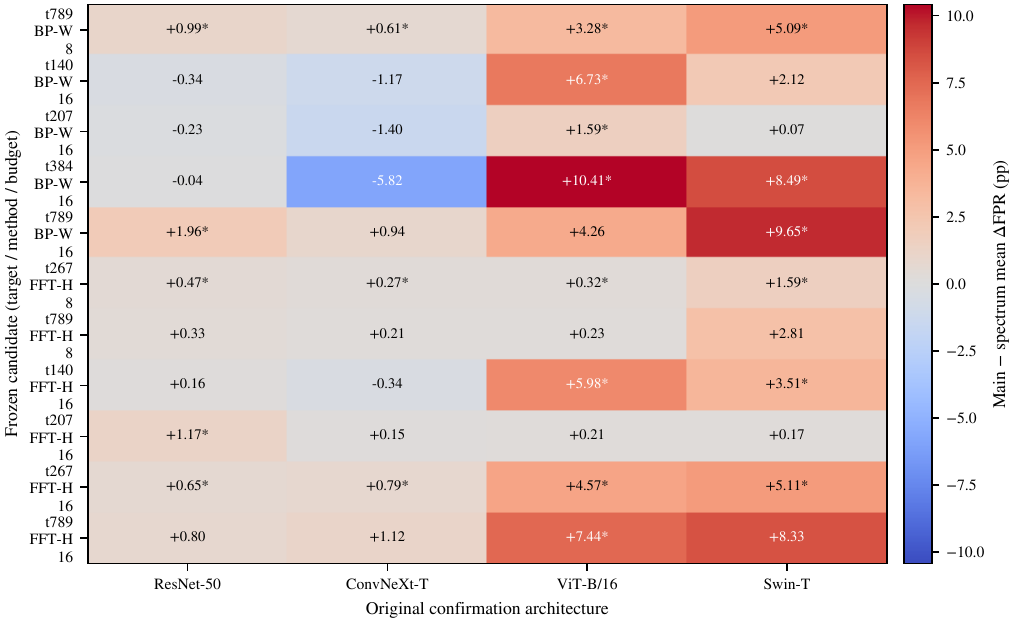}
\caption{Cellwise proposed-minus-spectrum target-FPR residuals in percentage points on the original four architectures. Asterisks mark Benjamini--Hochberg-corrected $q\leq.05$.}
\label{fig:spectrum-residual}
\end{figure*}

\begin{figure}[t]
\centering
\includegraphics[width=0.96\linewidth]{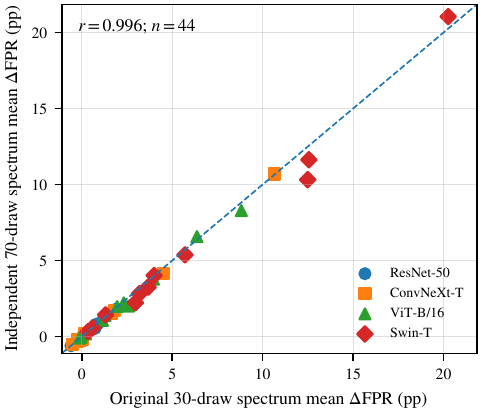}
\caption{Agreement between the original 30-draw and independent 70-draw spectrum-null means. The high agreement ($r=0.996$ across 44 cells) shows that extending the null primarily improves empirical-tail resolution rather than replacing it with a qualitatively different distribution.}
\label{fig:spectrum-null-stability}
\end{figure}

The spectrum-matched null reproduces a substantial portion of the mean effect, but the residual varies strongly by architecture and candidate. Table~\ref{tab:spectrum-by-model} summarizes the architecture-level result, Figure~\ref{fig:spectrum-residual} shows the cellwise residual, and Figure~\ref{fig:spectrum-null-stability} shows the agreement between the original 30-draw and independent 70-draw null estimates.

\begin{table}[t]
\centering
\small
\setlength{\tabcolsep}{2.5pt}
{
\caption{Residual target-FPR movement beyond 100 spectrum-matched random controls on the original architectures. Values are mean percentage-point changes.}
\label{tab:spectrum-by-model}
\begin{tabular}{@{}lrrrr@{}}
\toprule
Model & Proposed & Spectrum & Residual & $q\leq.05$ \\
\midrule
ResNet-50 & +0.907 & +0.369 & +0.539 & 5/11 \\
ConvNeXt-T & +1.112 & +1.534 & $-0.422$ & 3/11 \\
ViT-B/16 & +6.610 & +2.518 & +4.092 & 8/11 \\
Swin-T & +10.106 & +5.840 & +4.267 & 6/11 \\
\midrule
All cells & +4.684 & +2.565 & +2.119 & 22/44 \\
\bottomrule
\end{tabular}}
\end{table}

\begin{table}[t]
\centering
\small
\setlength{\tabcolsep}{3.1pt}
{
\caption{Spectrum-relative results for primary and secondary endpoints on the 44 original model--candidate cells.}
\label{tab:spectrum-secondary}
\begin{tabular}{@{}lrr@{}}
\toprule
Endpoint & Main $>$ spectrum mean & $q\leq.05$ \\
\midrule
Target FPR at 5\% calibration & 37/44 & 22/44 \\
Target-logit shift & 35/44 & 29/44 \\
Target-rank improvement & 37/44 & 31/44 \\
Targeted ASR@5 change & 40/44 & 21/44 \\
Targeted ASR@1 change & 13/44 & 0/44 \\
\bottomrule
\end{tabular}}
\end{table}

Spectrum-random controls are not evaluated on RegNet-Y-8GF or PVT-v2-B2. The held-out experiment therefore establishes transfer of the principal source-statistical effect, while the spectrum-relative residual is established only on the original four architectures.
\FloatBarrier

\section{\texorpdfstring{Reproducibility and Execution Environment}{Reproducibility and Execution Environment}}
\label{app:reproducibility}

\paragraph{Environment pinning.}
The experiments use the Nix flake's \texttt{x86\_64-linux} shell, which pins \texttt{nixpkgs} to revision \texttt{9ae611a455b9}, with the full revision recorded in \texttt{flake.lock}. Nix supplies Python 3.12 and system tools, while \texttt{uv} creates the local \texttt{.venv}. Core versions are PyTorch \texttt{2.7.0+cu126}, torchvision \texttt{0.22.0+cu126}, timm \texttt{1.0.28}, LIME \texttt{0.2.0.1}, and scikit-image \texttt{0.26.0}; \texttt{uv.lock} records the transitive environment.

\paragraph{Hardware and CUDA.}
The confirmation and revision runs used devices 0 and 1 of an eight-GPU Linux server with NVIDIA Tesla V100-SXM2 32~GB GPUs. The revision host used NVIDIA driver \texttt{580.126.20} (driver-reported CUDA compatibility 13.0), while PyTorch and torchvision use CUDA 12.6 wheels. The Nix shell exposes the host \texttt{libcuda} and NVML libraries; no custom CUDA extensions are compiled.

\paragraph{Models and evaluation.}
The code loads torchvision's default ImageNet-1K weights for ResNet-50, ConvNeXt-Tiny, ViT-B/16, Swin-T, and RegNet-Y-8GF, and loads PVT-v2-B2 with \texttt{timm.create\_model("pvt\_v2\_b2", pretrained=True)}. All architectures use the shared 224-pixel preprocessing. Models are frozen, placed in evaluation mode, and run under \texttt{torch.no\_grad()} with batch size 96 and eight data-loader workers; no training or fine-tuning is performed.

\paragraph{Reproduction entry point.}
Reproduction uses \texttt{nix develop} followed by \texttt{uv sync --extra dev}; \texttt{check\_cuda.py} and \texttt{pytest} provide CUDA and test preflights. The \texttt{run\_publication\_pipeline.sh} wrapper executes the staged pipeline end-to-end and records each stage's configuration.
\FloatBarrier

%% file: tbl/core_contributions_full.tex

\providecommand{\mixed}{\(\triangle\)}

\begin{table*}[!p]
\centering
\caption{
Comparison of attack-construction assumptions and contribution scope.
``Class-conditioned'' means that a target label \(y\) is selected before
constructing the attack signal, rather than assigned post hoc according
to the label to which predictions happen to collapse.
}
\label{tab:ours-but-not-theirs}

\begingroup
\small
\renewcommand{\arraystretch}{1.12}

\def\coretabrowrule{%
  \specialrule{0.20pt}{0.95pt}{0.95pt}%
}

\begin{tabularx}{\textwidth}{@{}
    >{\raggedright\arraybackslash}p{0.235\textwidth}
    @{\hspace{0.55pt}}
    >{\centering\arraybackslash}X
    *{8}{
      @{\hspace{0.55pt}}
      >{\centering\arraybackslash}X
    }
@{}}

\toprule

\textbf{Property}
&
\shortstack{
  \textbf{Fast}\\
  \textbf{Feature}\\
  \textbf{Fool}\\[-0.4ex]
  {\scriptsize\cite{Mopuri2017}}
}
&
\shortstack{
  \textbf{GAP}\\[-0.4ex]
  {\scriptsize\cite{Poursaeed2018}}
}
&
\shortstack{
  \textbf{Li}\\
  \textbf{et al.}\\[-0.4ex]
  {\scriptsize\cite{Li2020}}
}
&
\shortstack{
  \textbf{HIT}\\[-0.4ex]
  {\scriptsize\cite{Zhang2022}}
}
&
\shortstack{
  \textbf{Wenger}\\
  \textbf{et al.}\\[-0.4ex]
  {\scriptsize\cite{wenger2022naturally}}
}
&
\shortstack{
  \textbf{Han}\\
  \textbf{et al.}\\[-0.4ex]
  {\scriptsize\cite{Han2024NaturalTriggers}}
}
&
\shortstack{
  \textbf{Neuhaus}\\
  \textbf{et al.}\\[-0.4ex]
  {\scriptsize\cite{Neuhaus_2023_ICCV}}
}
&
\shortstack{
  \textbf{Texture}\\
  \textbf{Adv.}\\[-0.4ex]
  {\scriptsize\cite{Mou2024}}
}
&
\textbf{Ours}
\\

\midrule

No victim-model queries
&
\na
&
\na
&
\yes
&
\yes
&
\yes
&
\yes
&
\na
&
\yes
&
\yes
\\

\coretabrowrule

No victim gradients or internal representations
&
\no
&
\no
&
\yes
&
\yes
&
\yes
&
\yes
&
\no
&
\yes
&
\yes
\\

\coretabrowrule

No victim or surrogate model used during construction
&
\no
&
\no
&
\no
&
\yes
&
\yes
&
\mixed
&
\no
&
\yes
&
\yes
\\

\coretabrowrule

No learned attack model or image generator
&
\yes
&
\no
&
\no
&
\yes
&
\yes
&
\no
&
\yes
&
\yes
&
\yes
\\

\coretabrowrule

No hand-designed generic texture or geometric pattern
as the main attack signal
&
\yes
&
\yes
&
\yes
&
\no
&
\yes
&
\yes
&
\yes
&
\no
&
\yes
\\

\coretabrowrule

Uses labeled-dataset structure without a model
during trigger discovery
&
\no
&
\no
&
\no
&
\no
&
\yes
&
\mixed
&
\no
&
\no
&
\yes
\\

\coretabrowrule

No training-time poisoning, relabeling, or
attack-specific target-model retraining
&
\yes
&
\yes
&
\yes
&
\yes
&
\no
&
\no
&
\yes
&
\yes
&
\yes
\\

\coretabrowrule

Constructs a perturbation applicable to arbitrary
held-out source images
&
\yes
&
\yes
&
\yes
&
\yes
&
\no
&
\no
&
\no
&
\yes
&
\yes
\\

\coretabrowrule

Produces a perturbation indexed by a preselected
target label \(y\)
&
\no
&
\yes
&
\no
&
\no
&
\no
&
\yes
&
\no
&
\no
&
\yes
\\

\coretabrowrule

Test-time perturbation construction is simultaneously
model-blind and class-conditioned
&
\no
&
\no
&
\no
&
\no
&
\no
&
\no
&
\no
&
\no
&
\yes
\\

\coretabrowrule

Primary outcome is perturbation-induced
target-class false-positive inflation
&
\no
&
\no
&
\no
&
\no
&
\no
&
\no
&
\no
&
\no
&
\yes
\\

\coretabrowrule

Central claim: dataset-only construction reveals
shared target-specific vulnerabilities across
independently trained clean models
&
\no
&
\no
&
\no
&
\no
&
\no
&
\no
&
\no
&
\no
&
\yes
\\

\bottomrule

\end{tabularx}

\vspace{3pt}

\parbox{\textwidth}{
\footnotesize
\textit{Notes.}
\yes\ denotes yes, \no\ denotes no, \na\ denotes not applicable,
and \mixed\ denotes yes for some variants and no for others in papers that present multiple methods.
Wenger et al. identify physical trigger--class subsets from labeled
co-occurrence structure, but establish the backdoor by relabeling
co-occurring images and training a poisoned model.
Han et al.'s optimization-based trigger uses a substitute detector,
whereas their custom-trigger variant uses long-tailed facial attributes
without detector supervision; both variants use learned generators and
poisoned training samples.
Neuhaus et al. present a model-based spurious-feature discovery,
benchmarking, and mitigation framework rather than a perturbation method.
}

\endgroup
\end{table*}